\def\year{2022}\relax
\documentclass[letterpaper]{article} 
\usepackage{aaai22}  
\usepackage{times}  
\usepackage{helvet}  
\usepackage{courier}  
\usepackage[hyphens]{url}  
\usepackage{graphicx} 
\usepackage{natbib}  
\usepackage{caption} 
\DeclareCaptionStyle{ruled}{labelfont=normalfont,labelsep=colon,strut=off} 
\usepackage{algorithm}
\usepackage{algorithmic}
\usepackage{amsmath}
\usepackage{float}
\usepackage{graphicx}
\usepackage{subfig}
\usepackage{bbm}
\usepackage{multirow}
\usepackage{multicol}
\usepackage{array}
\usepackage{verbatim}
\usepackage{booktabs}       
\usepackage{hyperref}       
\usepackage{url}            
\usepackage{tablefootnote}
\usepackage{amsfonts} 

\usepackage{nicefrac}       
\usepackage{microtype}      
\usepackage{adjustbox}
\usepackage{tabularx}

\PassOptionsToPackage{numbers}{natbib}
\usepackage[dvipsnames]{xcolor}

\newcommand{\ba}{\mathbf{a}}

\newcommand{\be}{\mathbf{e}}
\newcommand{\bc}{\mathbf{c}}
\newcommand{\bh}{\mathbf{h}}

\newcommand{\bA}{\mathbf{A}}
\newcommand{\bW}{\mathbf{W}}

\newcommand{\br}{\mathbf{r}}

\newcommand{\mE}{\mathcal{E}}
\newcommand{\mG}{\mathcal{G}}
\newcommand{\mH}{\mathcal{H}}

\newcommand{\mR}{\mathcal{R}}

\newcommand{\mS}{\mathcal{S}}
\newcommand{\mA}{\mathcal{A}}

\newcommand{\one}{\mathbbm{1}}
\newcommand{\zero}{\mathbf{0}}
\newcommand{\Real}{\mathbbm{R}}

\newcommand{\LSTM}{\text{LSTM}}
\newcommand{\softmax}[1]{\sigma(#1)}
\newcommand{\relu}[1]{\text{ReLU}(#1)}

\newcommand{\tworow}[1]{\multirow{2}{*}{#1}}

\newcommand{\highest}[1]{\textbf{#1}}

\newcommand{\tb}{\textbf}
\newcommand{\ti}{\textit}

\newcommand{\ts}{\textsc}

\newcommand{\model}{CURL}

\newcommand{\mc}{\mathcal}

\newcommand{\prev}[1]{}

\usepackage{newfloat}
\usepackage{listings}
\floatstyle{ruled}
\newfloat{listing}{tb}{lst}{}
\floatname{listing}{Listing}
\title{Learning to Walk with Dual Agents for Knowledge Graph Reasoning}

\author{Denghui Zhang$^{1}$$^{\dagger}$, Zixuan Yuan$^{1}$$^{\dagger}$, Hao Liu$^{2}$, Xiaodong Lin$^{1}$, Hui Xiong$^{1}$\thanks{
Hui Xiong is the corresponding author.\\
$^{\dagger}$~Equal contribution. Author ordering determined by coin flip.}}
\affiliations{$^{1}$Rutgers University\\ $^{2}$Hong Kong University of Science and Technology (HKUST)\\
\{denghui.zhang, zy101, lin, hxiong\}@rutgers.edu, liuh@ust.hk}

\usepackage{bibentry}

\begin{document}

\maketitle

\begin{abstract}
Graph walking based on reinforcement learning (RL) has shown great success in navigating an agent to automatically complete various reasoning tasks over an incomplete knowledge graph (KG) by exploring multi-hop relational paths.
However, existing multi-hop reasoning approaches only work well on short reasoning paths and tend to miss the target entity with the increasing path length.
This is undesirable for many reasoning tasks in real-world scenarios, 
where short paths connecting the source and target entities are not available in incomplete KGs,  
and thus the reasoning performances drop drastically unless the agent is able to seek out more clues from longer paths. 
To address the above challenge, in this paper, we propose a dual-agent reinforcement learning framework, which trains two agents (\ts{giant} and \ts{dwarf}) to walk over a KG \ti{jointly} and search for the answer \ti{collaboratively}.
Our approach tackles the  reasoning challenge in long paths by assigning one of the agents (\ts{giant}) searching on cluster-level paths quickly and providing stage-wise hints for another agent (\ts{dwarf}). 
Finally, experimental results on several KG reasoning benchmarks show that our approach can search answers more accurately and efficiently, and outperforms existing RL-based methods for long path queries by a large margin.
\end{abstract}

\section{Introduction}
\begin{figure*}[t]
	\centering
	\includegraphics[width=.99\linewidth]{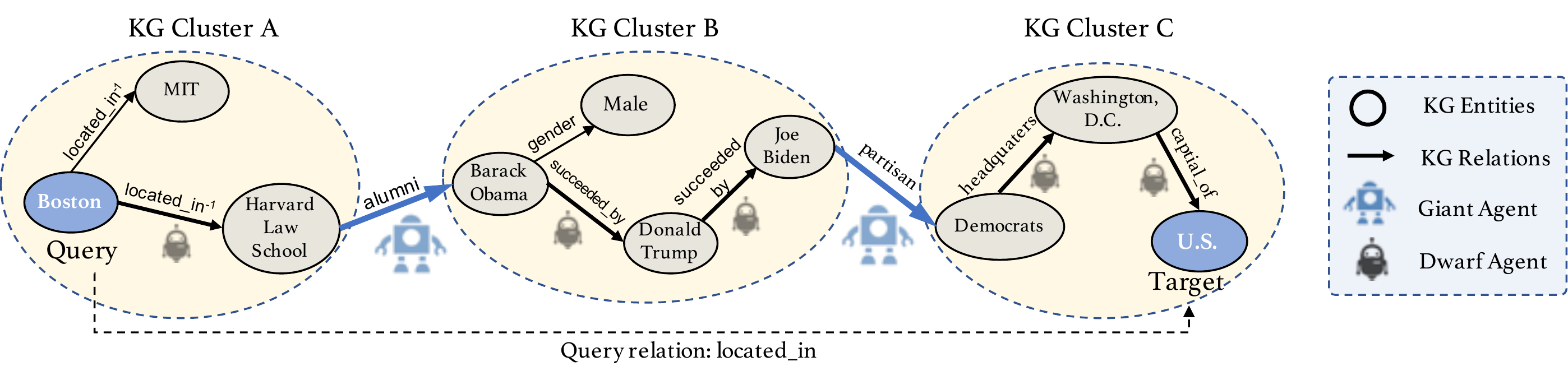}
		\vspace{-5pt}
	\caption{An illustrative diagram of the dual-agent reinforcement walking approach.
	Two agents work collaboratively to find the target answer (\ts{giant} walks by clusters, \ts{dwarf} walks by entities). 
    }
	\label{fig:approach}
	\vspace{-10pt}
\end{figure*}

Knowledge graphs (KGs) \prev{support a variety of knowledge-driven applications} have become an essential building block of various knowledge-driven services, such as relation extraction \cite{mintz2009distant}, question answering \cite{cui2019kbqa}, and recommender systems \cite{zhang2016collaborative}.
A KG is usually defined as a directed graph $\mG = (\mathcal{E}, \mathcal{R})$, where $\mathcal{E}$ is a collection of entity nodes, and $\mathcal{R}$ is a set of relation edges.
Due to the highly incomplete nature, in practice, KGs often fail to include sufficient fact triples to satisfy the long-tail scenarios in various tasks.
To this end, we focus our study in the context of automatic KG reasoning, also known as knowledge graph completion (KGC), i.e., constructing $f(e_s, r_q, ?|\mG)$ or $f(?, r_q, e_t|\mG)$ to infer missing facts by synthesizing information from multi-hop paths between the source and target nodes.
One example is illustrated in Figure \ref{fig:approach}, where no direct link can be found on between the target node $e_t=$ "U.S." and the source node $e_s=$ "Boston".
However, by leveraging existing indirect links and the query relation $r_q=$ "LocatedIn", one is possible to infer the fact (\ts{Boston}, \ts{LocatedIn}, \ts{U.S.}).

\prev{Recently, extensive research has emerged on learning latent vectors of entities and relations for knowledge graph reasoning using tensor factorization or neural networks \cite{wang2014knowledge, yang2014embedding, trouillon2016complex}. 
Unfortunately, these embedding-based approaches focus on short-term structural information and primarily consider single-hop reasoning, lacking the expressibility for modeling long chains of reasoning in KGs.}

In the past several years, extensive research has been conducted on learning latent representations of entities ($e \in \mathcal{E}$) and relations ($r \in \mathcal{R}$) for knowledge graph reasoning by using tensor factorization or neural networks \cite{wang2014knowledge, yang2014embedding, trouillon2016complex}.
Such embedding-based approaches mainly focus on preserving the structural information in the KG and are effective for single-hop reasoning.
Also, recent works have considered exploiting reinforcement learning (RL) \cite{das2017go, xiong2017deeppath, shen2018m} for KGC reasoning tasks, where a walking agent is leveraged over KG paths to compose single-hop triplets into multi-hop reasoning chains. 
For instance, MINERVA \cite{das2017go} is an end-to-end model that adopts REINFORCE algorithm \cite{sutton1999policy} to train the RL agent to search over KGs starting from the source and arrive at the candidate answers.

However, a noteworthy issue of these walking-based models is that they rely heavily on short reasoning chains (e.g., {maximum\_path\_length=3} in MINERVA), where the performance drops drastically if short indirect paths are also absent. 
Indeed, such single-agent approaches often get stuck when reasoning on a long path.
The reasons are two-fold. 
First, KGs consist of massive entities and relations, the dimension of the discrete action space at each step is typically large \cite{das2017go}. 
As a result, the difficulty of reasoning (i.e., making right decisions constantly) increases drastically with the increasing number of reasoning steps.
Without narrowing down the scope of representative entities and relations, the underlying agent may conduct unnecessary traverse among similar objects, and thus has low efficiency for path finding.
Second, prior approaches train the agent with sparse rewards. 
Specifically, they only return a positive reward when the agent reaches the target entity by the end of a walking, and penalizes all actions within the path otherwise.
This may result in
false-negative rewards to the intermediate actions which are reasonable, and hinders the policy network from learning trustworthy long-term patterns.



To tackle the above problems, in this paper, we propose a \tb{C}ollaborative D\tb{u}al-agent \tb{R}einforcement \tb{L}earning framework, named \model. 
Unlike existing walking-based RL models, which rely on one single agent to explore reasoning paths over entities and relations in KGs, we design two agents, \ts{Giant} agent and \ts{Dwarf} agent, to perform reasoning at different granularities and search for the answer collaboratively.
Specifically, \ts{Giant} performs coarse-grained reasoning by walking rapidly over pre-defined abstractive KG clusters, while \ts{Dwarf} carefully traverses entities within each cluster to perform fine-grained reasoning.
By leveraging a \emph{Collaborative Policy Network},
two agents can share historical path information with each other to enhance their state representations.
Moreover, we propose a \emph{Mutual Reinforcement Reward System} to overcome the sparse reward issue. 
Instead of using a static final reward, we allow \ts{dwarf} to borrow weighted reward from \ts{giant} for its intermediate steps and vice versa.
Intuitively, \ts{giant} provides abstract and milestone-like hints to guide \ts{dwarf}'s behavior and reduce the search space.
On the other hand, \ts{dwarf} provides regularization over \ts{giant}'s behavior to help avoid less informative cluster-level reasoning chains.
By training two agents jointly, 
our framework exploits the KG structure more thoroughly, i.e., from both global and local views, long and short reasoning paths, macro and micro trajectories, to improve the effectiveness of KG reasoning tasks.
We conduct extensive experimental studies on three real-world KG datasets. 
The results demonstrate that our approach can search answers more accurately and efficiently than existing embedding-based approaches as well as traditional RL-based methods, and outperforms them on long path queries significantly.




\section{Methodology}
In this section, we first review the formal problem definition of knowledge graph reasoning and the single-agent reinforcement learning approach by \citep{das2017go}.
Then we introduce our dual-agent based approach that explores path patterns at two granularity levels to tackle the long-path challenge and sparse reward problem.

\subsection{Problem Definition}
Formally, a knowledge graph $\mG$ is represented as a directed graph: $\mG = \{(e_s, r, e_o), e_s,e_o\in\mE, r\in \mR\}$, where $\mE$ is the set of entities and $\mR$ is the set of relations. 
Each directed link in the knowledge graph $l = (e_s, r, e_o)\in\mG$ corresponds to a real-world fact tuple, e.g., \ts{(Joe Biden, president\_of, United States}). 
For the knowledge graph reasoning task, we follow the definition in prior graph walking literature, a.k.a, query answering task \cite{das2017go}.
Specifically,
given a query $(e_s, r_q, ?)$, where $e_s$ is the source entity and $r_q$ is the relation of interest,
the goal of query answering is to perform efficient search over $\mG$ and find the possible answers (i.e., correct target entities) $E_o=\{e_o\}$ where $(e_s, r_q, e_o)\notin\mG$ due to incompleteness of KG. 

Although being a promising way to discover new fact knowledge, this task is challenging as it requires an algorithm to be capable of sophisticated multi-hop reasoning over the incomplete KG.
\prev{Note that we follow the same \ti{end-to-end walking for reasoning} paradigm as MINERVA in this work, mainly because of its desirable properties such as eliminating the need to pre-train agent or pre-compute path features, avoiding ranking all entities in the graph.}
Note that we follow the \ti{end-to-end walking for reasoning} paradigm as in MINERVA, to avoid undesirable requirements such as agent pre-training, path features pre-computing, and all entities ranking in the graph.

\subsection{Previous Single-Agent Reinforcement Walking}
\label{subsec:minerva}
The process of walking (searching) on KG can be viewed as a Markov Decision Process (MDP)~\citep{sutton2018reinforcement}:
given the query entity $e_s$ and relation $r$, the agent departs from $e_s$,
sequentially selects an outgoing edge $l$ and traverses to a new entity until it arrives at a target answer or reaches maximum path length.
The MDP can be expressed by the following essential components (we eliminate \ts{Observation} part in MINERVA \citep{das2017go} for simplicity, yet the formulation is equivalent).




\paragraph{State, Action, Transition, Reward}
Each state $s_t=(e_t, (e_s, r_q))\in\mS$ is a tuple where $e_t$ is the entity visited at step $t$ and $(e_s, r_q)$ are the source entity and query relation. 
$e_t$ can be viewed as state-dependent information while $(e_s, r_q)$ are the global context shared by all states.
The set of possible actions $A_t\in\mA$ of at step $t$ consists of the outgoing edges of $e_t$ in $\mG$. Concretely, $A_t=\{(r', e') | (e_t, r', e')\in\mG\}$. 
To grant the agent an option to terminate a search, a self-loop edge is added to every $A_t$. 
Because search is unrolled for a fixed number of steps $T$, the self-loop acts similarly to a ``stop'' action.
A transition function $\delta: \mS\times\mA\rightarrow\mS$ is defined by $\delta(s_t, A_t)=\delta(e_t, (e_s, r_q), A_t)$. 
Action probability is predicted by a policy network $\pi_{\theta}$, which takes as input the state information.
Popular choices for $\pi_{\theta}$ includes simple models like MLP and sequence-aware models like RNN.
In the default formulation, the agent receives a terminal reward of 1 if it arrives at a correct target entity at the end of search (i.e., within maximal steps) and 0 otherwise, i.e., $R_b(s_T) = \one\{(e_s, r_q, e_T)\in\mG\}$.


\subsection{Our Dual-Agent Reinforcement Walking}

The above single-agent approach can effectively explore short paths on KG and discover short chains of reasoning. However, when the path length increases, the agent tends to miss the target entity and fails to catch meaningful long chains of reasoning.
Contrarily, our approach launches two agents: \ts{Giant Agent} and \ts{Dwarf Agent} (short as \ts{Giant} and \ts{Dwarf} respectively), to collaboratively explore paths at different granularity levels and search for the answer.
\ts{Giant} walks rapidly over inner clusters of the KG, \ts{Dwarf} slowly traverses by entities inside the clusters, while meantime, they share essential path and reward information to each other, taking advantage of a more comprehensive view (i.e., both cluster/global view and entity/local view) of KG to enhance reasoning.
Figure \ref{fig:approach} presents a concise illustration of our approach CURL. 

\paragraph{Mapping KG to Clusters} 
We first divide an original KG into $N$ clusters of nodes using K-means \cite{macqueen1967some} on the pre-trained entity embeddings\footnote{We apply TransE \citep{bordes2013translating} for its efficiency in encoding the structural closeness information of KG.}. 
Based on these clusters, we also build a cluster-wise connection graph $\mG^{c}$,
where two clusters will be connected if there is at least one entity-level edge between them.
It can be viewed as a denser mapping of the original KG.
\ts{giant} aims to walk over $\mG^{c}$ to reach a ``fuzzy answer'', i.e., the target cluster in which the end entity lies.

\paragraph{\ts{giant agent}: Cluster-Level Exploration}
Following the RL paradigm, \ts{giant} makes moves based on the state.
Each state $s_t^c=(c_t, c_s)\in\mS^c$ is a tuple where $c_t\in\mG^c$ represents the cluster visited at step $t$ and $c_s$ is the start cluster which the source entity belongs to.
Specifically, $c_t$ contains state-dependent information while $c_s$ are the global context shared by all states.
At each step $t$, the possible actions for \ts{giant} consists of neighbor clusters of $c_t$ in $\mG^c$.
Concretely, $A_t^c=\{c' | (c_t, c')\in\mG^c\}$.
In other words, it traverses the KG in a fashion of cluster by cluster.
Since cluster-level paths are normally shorter than entity-level paths (e.g., in Figure \ref{fig:approach}, $\text{length}_{cluster}=3$, $\text{length}_{entity}=7$), 
\ts{Giant} should be allowed to "stay" in some clusters during walking in order to \ti{synchronize} with \ts{Dwarf}.
To fulfill this purpose, a special action, i.e., \ts{stop}, is also added into every action space  $A_t^c$.
Each action $a_t^c\in A_t^c$ is made based on the prediction of the policy network of \ts{Giant}: $\pi_{\theta}^c$.
After walking multiple required steps, \ts{giant} receives a terminal reward of 1 if it arrives at the cluster $c_T$ where a correct entity answer lies in $e_T\in c_T$, and 0 otherwise.

\paragraph{\ts{dwarf agent}: Entity-Level Exploration}
Similar to the single-agent approach,
\ts{dwarf} agent walks over the original KG $\mG$ to reach an accurate answer, i.e., a target entity.
Specifically, each state $s_t^e=(e_t, (e_s, r_q))$ is a tuple consisting of the current entity being visited $e_t$, and the query entity, relation, $e_s$ and $r_q$.
At each step $t$, to make an action, $\ts{dwarf}$ selects one from all the outgoing edges of the current entity, $A_t^e=\{(r', e') | (e_t, r', e')\in\mG\}$.
The selection is predicted by its own policy network $\pi_\theta^e$.
Within a maximum number of steps, if the agent arrives at a correct target entity, it will receive a final reward of 1 and 0 otherwise.




\subsubsection{Collaborative Graph Walking: Jointly Train $\pi_{\theta}^e$ and $\pi_{\theta}^c$}
Under the framework of dual-agent, to maximize the benefit of both sides, i.e., cluster-level and entity-level exploration, 
we propose two advances to tame the two distinct agents to walk in a mutually beneficial way, 
namely, 
Collaborative Policy Networks and Mutual Reinforcement Rewards.

\paragraph{Collaborative Policy Networks} 
Agents make moves based on the output of policy network, 
we use two separate networks $\pi_{\theta}^e$ and $\pi_{\theta}^c$ respectively\footnote{The superscript $c,e$ stands for cluster/entity.} to model the action selection of \ts{giant} and \ts{dwarf}.
Specifically, every entity, relation in $\mG$, as well as cluster in $\mG^c$ is assigned a dense vector embedding $\be\in\Real^d$, $\br\in\Real^d$, $\bc\in\Real^{2d}$. 
We use these embeddings to represent RL actions and states of two agents.
For \ts{dwarf}, each action $a_t^e=(r_{t}, e_{t})\in A_t^e$ consisting of the next outgoing relation and entity, is represented as the concatenation of the relation embedding and the end node embedding $\ba_t^e = [\br_t; \be_t]\in\Real^{2d}$.
For \ts{giant}, the action corresponds to the next outgoing cluster and we directly use the cluster embedding to represent it, i.e., $\ba_t^c = \bc_t\in\Real^{2d}$.
For both $\pi_{\theta}^e$ and $\pi_{\theta}^c$,
we first use two separate LSTMs to encode their search history $h_t^c= (c_1, \dots, c_t)\in\mH^c$, $h_t^e= (e_s, r_1, e_1, \dots, r_t, e_t)\in\mH^e$ according to the recurrent dynamics,
\begin{align}
\label{eq:history}
&\bh_0^c = \LSTM_c(\zero, \bc_0), \	\bh_0^e = \LSTM_e(\zero, [\br_0; \be_s]) ,\\
&\bh_{t}^{c}  = \LSTM_c(\bW^c[\bh_{t-1}^c;\bh_{t-1}^e], \ \ba_{t-1}^c), \ t > 0, \\
&\bh_t^{e}  = \LSTM_e(\bW^e[\bh_{t-1}^e;\bh_{t-1}^c],\  \ba_{t-1}^e), \ t > 0, 
\end{align}
where we modify their internal structures to allow \ti{state sharing} between $\LSTM_c$ of \ts{giant} and $\LSTM_e$ of \ts{dwarf}.
Specifically, at each step $t>0$, 
we calculate the concatenation of the two raw states $\bh_{t}^c,\bh_{t}^e\in\Real^{2d}$ as the new states $[\bh_{t}^c;\bh_{t}^e], [\bh_{t}^e;\bh_{t}^c]\in\Real^{4d}$. 
In addition, we apply two transforming matrices $\bW^c,\bW^e \in \Real^{2d\times 4d}$ to reduce the dimension of the new states,
otherwise, their dimension will increase exponentially w.r.t the number of steps.
With the modification,
each hidden state $\bh_{t}^{c}$ ($\bh_{t}^{e}$) of an agent is conditioned on: its own previous state $\bh_{t-1}^{c}$ ($\bh_{t-1}^{e}$), the other agent's previous state $\bh_{t-1}^{e}$ ($\bh_{t-1}^{c}$), its previous action $\ba_{t-1}^{c}$ ($\ba_{t-1}^{e}$).
This ensures sharing essential path information between \ts{giant} and \ts{dwarf}, as to some extent, cluster-level are complementary to entity-level paths, it can consequently enable more robust action selection for both of them.

To predict the next action (i.e., the next cluster for \ts{giant} and next relation-entity edge for \ts{dwarf}), we further apply a two-layer feedforward network with ReLU nonlinearity on the concatenation of their last LSTM states and current RL state embeddings,
\begin{align}
    \label{eq:mlp1}
	\pi_{\theta}^c(a_t^c|s_t^c) &= \softmax{\bA_t^c \times \bW_2^c~\relu{\bW_1^c[\bc_t;\bh_t^c]}}, \\
	\label{eq:mlp2}
	\pi_{\theta}^e(a_t^e|s_t^e) &= \softmax{\bA_t^e \times \bW_2^e~\relu{\bW_1^e[\be_t;\br_q;\bh_t^e]}},
\end{align}
where $\bW_1^c,\bW_2^c\in\Real^{4d\times4d},\bW_1^e,\bW_2^e \in\Real^{6d\times6d}$ are the matrices of learnable weights, $\bA_t^c \in\Real^{|A_t^c|\times4d},\bA_t^e\in\Real^{|A_t^e|\times6d}$ represent the embeddings of all next possible actions for \ts{giant} and \ts{dwarf}.
$\sigma$ denotes the softmax operator. 

\paragraph{Mutual Reinforcement Rewards} 
The default rewards for \ts{giant} and \ts{dwarf} only consider whether their own agent arrives at the target cluster or entity, 
leaving two problems limiting the dual agent to collaborate effectively:
(i) As \ts{giant} walks over the cluster graph which is more densely connected, it leads to more diverse trajectories and
hard to be consistent with entity-level paths.
(ii) \ts{dwarf} does not learn any stage-wise knowledge from \ts{giant} explicitly.
To alleviate the issue, we provide a new mutual reinforcement reward system, which can
(1) constrain \ts{giant} to generate cluster trajectories consistent with entity-level paths;
(2) allow \ts{giant} to provide stage-wise hints for \ts{dwarf} to follow. 
Specifically, both agents' final reward consists of two parts, i.e., their own default reward and an auxiliary weighted reward borrowed from their partner, 
\begin{align}
	\label{eq:mreward1}
    R_c(s_t^c) = \underbrace{r_c(s_T^c)}\limits_{\ti{default reward}} + \underbrace{\Phi(s_t^c, s_t^e)\cdot r_e(s_T^e)}\limits_{\ti{partner reward}}, \ \ t\in[1,T],\\
    \label{eq:mreward2}
  R_e(s_t^e) = \underbrace{r_e(s_T^e)}\limits_{\ti{default reward}} + \underbrace{\Phi(s_t^e, s_t^c)\cdot r_c(s_T^c)}\limits_{\ti{partner reward}}, \ \ t\in[1,T],
\end{align}
where $r_c(s_T^c)$ and $r_e(s_T^e)$ denote the default final rewards for \ts{giant} and \ts{dwarft}, defined as $r_c(s_T^c) = \one\{\exists e_a\in c_T, (e_s, r_q, e_a)\in\mG\}, r_e(s_T^e) = \one\{(e_s, r_q, e_T)\in\mG\}$. 
Moreover, $\Phi(s_t^c, s_t^e)$ is an evaluation function measuring the consistency of each action made by two agents.
In practice, it is calculated as the cosine similarity between the pre-trained embeddings\footnote{Each cluster embedding is obtained by averaging all entity embeddings within it.} of the current traversed cluster and entity, i.e., $\Phi(s_t^c, s_t^e)=\frac{\bc_t^\top\be_t}{||\bc_t||_2||\be_t||_2}$.
For \ts{giant}, 
its partner reward will be valid only if 
\ts{dwarf} reaches the target entity by the end, and meantime, the current cluster visited must be close to the corresponding entity at step $t$ so that the weight is sufficient.
Similarly, for \ts{dwarf}, the partner reward will be valid only if \ts{giant} reaches the target cluster by the end and the consistency weight is sufficient.
This ensures that both agents learn knowledge from partner only at the right time, i.e., when their partner succeeds to reach the correct target.
The measuring coefficient $\Phi(s_t^c, s_t^e)$ controls the strength of partner reward based on the path consistency, i.e., the overlap between the cluster-level path and entity-level path.

\begin{algorithm}[ht]
    \small
	\begin{algorithmic}[1]
	\STATE {\bfseries Input: }{KG $\mc{G}$; Initial entity and cluster nodes $e_s$, $c_s$;  Entity-level query $r_q$;  Target entity and cluster nodes $e_{o}$, $c_{o}$; Maximum path length $T$; Episode size $P$, Rollout size $L$}
	\STATE {\bfseries Output:}{Well-trained policy networks $\pi^{c}_{\theta}$, $\pi^{e}_{\theta}$ of \ts{giant} and \ts{dwarf}}
	\FOR{episode $p$ in $\{1, \dots, P\}$}
		\STATE Set current entity and cluster nodes $e_0 = e_s$, $c_0 = c_s$ 
		\FOR{$t=0, \ldots, T-1$}
		
		\STATE Predict the next cluster $c_{t+1}$ for \ts{giant} and next relation-entity edge $(r_{t+1}, e_{t+1})$ for \ts{dwarf} based on Eq. (\ref{eq:mlp1}) - (\ref{eq:mlp2})\;
		\ENDFOR
		\STATE Set default cluster-level reward $r_c = 1$ if the end of the path $c_{T} = c_{o}$ otherwise $r_c=0$\; 
		\STATE Set default entity-level reward $r_e = 1$ if the end of the path $e_{T} = e_{o}$ otherwise $r_e=0$\; 
		\STATE Repeat lines 5 - 9 for running $L$ rollouts (see the expectation in Eq. (8) - (9)) to update the cluster-level and entity-level policies
		\STATE Compute the mutual reinforcement rewards $R_e(s^e_t)$, $R_c(s^c_t)$ based on Eq. (\ref{eq:mreward1}) - (\ref{eq:mreward2})
		\STATE Update the model parameters with REINFORCE:
		\begin{align}
		\theta^c	&\leftarrow	\theta^c  + \alpha \cdot \nabla_{\theta^c}
		\mathbb{E}_{A^{c}_1, \dots, A^{c}_T \sim \pi_{\theta}^c} 
        \sum\limits_{t=0}^{T-1}[R_c(s^c_t) | s^c_0] 
		\\
		\theta^e	&\leftarrow	\theta^e  + \alpha \cdot \nabla_{\theta^e} 
		\mathbb{E}_{A^{e}_1, \dots, {A}^{e}_T \sim \pi_{\theta}^e} \sum\limits_{t=0}^{T-1}[R_e(s^e_t) | s^e_0]
		\end{align}
	\ENDFOR	
		\RETURN $\pi_{\theta}^c, \pi_{\theta}^e$
	\caption{\model~Training Algorithm}
	\label{alg:reinforcewalk}
	\end{algorithmic}
\end{algorithm}
The detailed training procedure of \model~is described in Algorithm \ref{alg:reinforcewalk}. 
During inference, we use the same procedure of lines 3-7 below to calculate the action probabilities at each step. 
To find the target answer, we do a beam search with a beam width of 50 on \ts{dwarf} agent and rank entities by the probabilities of the trajectory that \ts{dwarf} took to reach the entity, all other unreachable entities get a rank of $\infty$.

\section{Experiments}

\subsection{Experiment Setup}
We evaluate the effectiveness of \model\footnote{Source code: https://github.com/RutgersDM/DKGR/tree/master}~by performing two fundamental tasks using three real-world KG datasets, i.e., FB15K-237, WN18RR, and NELL-995.
The WN18RR \cite{dettmers2018convolutional} and FB15K-237 \cite{toutanova2015representing} datasets are separately created from the original WN18 and FB15K datasets by removing various sources of test leakage, making the datasets more realistic and challenging.
The NELL-995 dataset released by \cite{xiong2017deeppath} contains separate graphs for each query relation. 
We compare against two classes of state-of-the-art KG reasoning baselines: KG representation learning methods (TransE \cite{wang2014knowledge}, TransR \cite{lin2015learning}, DistMult \cite{yang2014embedding}, CompIEx \cite{trouillon2016complex}) and multi-hop neural approaches (NeuralLP \cite{yang2017differentiable}, PRA \cite{lao2011random}, DeepPath \cite{xiong2017deeppath}, MINERVA \cite{das2017go}, M-Walk \cite{shen2018m}).
For reproducing all baseline results, we used the implementation released by authors on the best hyperparameter settings reported by them \footnote{We obtain and report the best results using the code and hyperparameters released by the authors of the baseline models.}. 
\model~and all baselines are implemented with Pytorch framework \cite{paszke2019pytorch} and run on a single 2080 Ti GPU. 
More experimental details and hyperparameters of our model are illustrated in Appendix A. 
According to \cite{shen2018m}, KG reasoning can be divided into the following tasks:

\begin{table}[t]
    \centering
    \huge
    \hspace*{-0.32cm}
    \setlength\tabcolsep{3.1pt}
    \setlength\extrarowheight{3pt}
    \scalebox{0.43}{
    \begin{tabular}{l|cccc|cccc|cccc}
        \toprule
    	\tworow{\textbf{Model}} &
    	\multicolumn{4}{c}{\textbf{FB15K-237}} & \multicolumn{4}{c}{\textbf{WN18RR}} & \multicolumn{4}{c}{\textbf{NELL-995}} \\
        \cline{2-13}
        & \huge{@1} & \huge{@3} & \huge{@10} & \huge{MRR} & \huge{@1} & \huge{@3} &  \huge{@10} & \huge{MRR} & \huge{@1} & \huge{@3} & \huge{@10} & \huge{MRR} \\
        \hline
        TransE & 24.8 & 40.1 & 45.0 & 36.1 & 28.9 & 46.4 & 53.4 & 35.9 & 51.4 & 67.8 & 75.1 & 45.6 
        \\
        DistMult & 27.5 & 41.7 & 56.8 & 37.0 & 41.0 & 44.1 & 47.5 & 43.3 & 61.0 & 73.3 & 79.5 & 68.0 \\
        ComplEx & \textbf{30.3} & \textbf{43.4} & \textbf{57.2} & \textbf{39.4} & 38.2 & 43.3 & 48.0 & 41.5 & 61.2 & 76.1 & 82.1 & 68.4
        \\
        \hline
        NeuralLP & 16.6 & 24.8 & 34.8 & 22.7 & 37.6 & 46.8 & \textbf{65.7} & 45.9 & -- & -- & -- & -- \\
        MINERVA & 19.2 & 30.7 & 42.6 & 27.1 & 41.3 & 45.6 & 51.3 & 44.8 & 58.8 & 74.6 & 81.3 & 67.5 \\
        M-Walk & 16.8 & 24.5 & 40.3 & 23.4 & 41.5 & 44.7 & 54.3 & 43.7 & 63.2 & 75.7 & 81.9 & 70.7 \\
        \model~ &  22.4 & 34.1& 47.0& 30.6 & \highest{42.9} & \highest{47.1} & 52.3 & \highest{46.0} & \textbf{66.7} & \textbf{78.6} & \textbf{84.3} & \highest{73.8}
        \\
        \bottomrule
    \end{tabular}
    }
    \caption{Query answering performance compared to state-of-the-art embedding based approaches (top part) and multi-hop reasoning approaches (bottom part). The Hits@1, 3, 10 and MRR metrics were multiplied by 100. We highlight the best approach in each category.
    \label{tb:mainresults}
    } 
\end{table}

\begin{figure*}[t]
\setlength{\abovecaptionskip}{0pt}
\setlength{\belowcaptionskip}{-10pt}
\centering
\subfloat[TeamPlaysSport.]{
\begin{minipage}[t]{0.21\textwidth}
\includegraphics[width=1\textwidth]{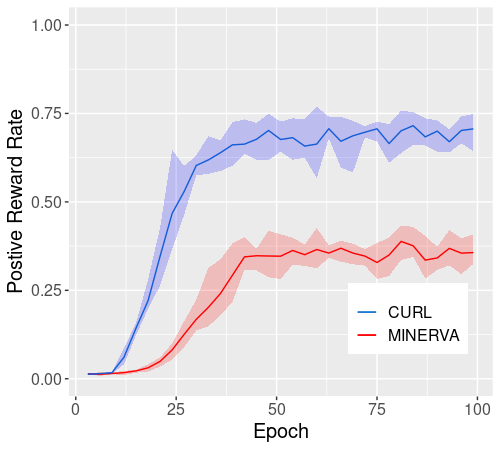}
\vspace{-10pt}
\label{ucncertain}
\end{minipage}%
}
\hspace{4pt}
\subfloat[AthletePlaysInLeague.]{
\begin{minipage}[t]{0.21\textwidth}
\includegraphics[width=1\textwidth]{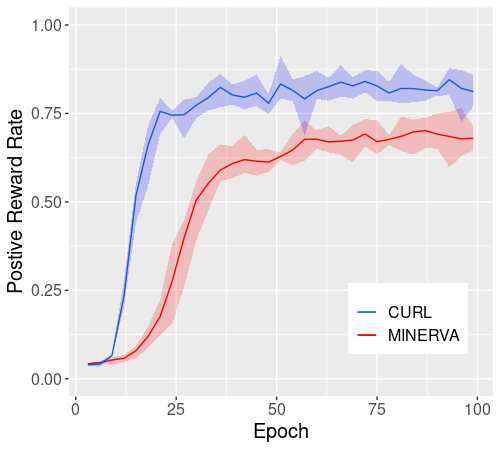}
\vspace{-10pt}
\label{geo}
\end{minipage}%
}
\hspace{4pt}
\subfloat[PersonBornInLocation.]{
\begin{minipage}[t]{0.21\textwidth}
\includegraphics[width=1\textwidth]{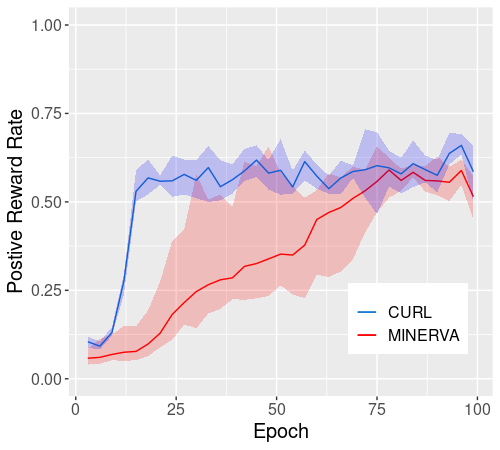}
\vspace{-10pt}
\label{time}
\end{minipage}%
}
\hspace{4pt}
\subfloat[NELL-995.]{
\begin{minipage}[t]{0.21\textwidth}
\includegraphics[width=1\textwidth]{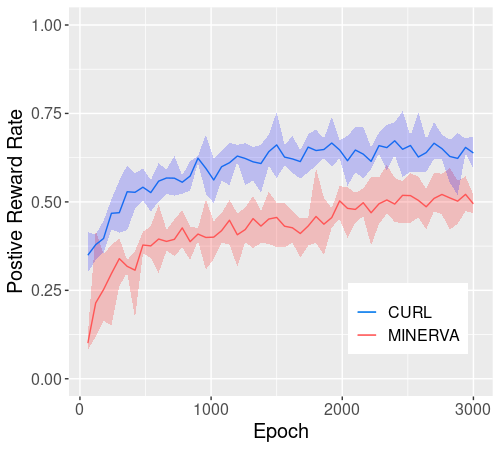}
\vspace{-10pt}
\label{user}
\end{minipage}%
}
\vspace{5pt}
\centering
\caption{ The positive reward rate on three NELL-995 relation tasks (a - c) and all tasks (d), and our agent is consistently better than MINERVA.} 
\label{positive_reward}
\end{figure*}
\begin{table}[t]
    \centering
    \huge
    \hspace*{-0.32cm}
    \setlength\tabcolsep{1.3pt}
    \setlength\extrarowheight{1pt}
    \scalebox{0.4}{
    \begin{tabular}{l|cc|ccccc}
    	\toprule
    	\textbf{Task} &
    	\textbf{TransR} &
    	\textbf{TransE} &
    	\textbf{PRA} & 
    	\textbf{DeepPath} & \textbf{MINERVA} & \textbf{M-Walk} & \textbf{\model}
        \\ \hline
        AthletePlaysInLeague & 91.2 & 77.3 & 84.1 & 96.0 & 94.0 & 96.1 & \highest{97.1}\\
        AthletePlaysForTeam & 67.3 & 62.7 & 54.7 & 75.0 & 80.0 & \highest{84.7} & 82.9\\
        AthleteHomeStadium & 72.2 & 71.8 & 85.9 & 89.0 & 89.8 & 91.9 & \highest{94.3}\\
        TeamPlaysSports & 81.4 & 76.1 & 79.1 & 73.8 & 88.0 & 88.4 & \highest{88.7}\\
        AthletePlaysSport & 96.3 & 87.6 & 47.4 & 95.7 & 98.0 & 98.3 & \highest{98.4}\\
        OrganizationHiredPerson & 73.7 & 71.9 & 59.9 & 74.2 & 85.6 & \highest{88.8} & 87.6\\
        PersonBornInLocation & 81.2 & 71.2 & 66.8 & 75.7 & 78.0 & 81.2 & \highest{82.1}\\
        WorksFor & 69.2 & 67.7 & 68.1 & 71.1 & 81.0 & \highest{83.2} & 82.1 \\
        OrgHeadquarteredInCity & 65.7 & 62.0 & 81.1 & 79.0 & 94.0 & 94.3 & \highest{94.8} \\
        PersonLeadsOrganization & 77.2 & 75.1 & 70.0 & 79.5 & 87.7 & 88.3 & \highest{88.9}\\ 
        \bottomrule
    \end{tabular}
    }
    \caption{Performance on fact prediction, MAP scores for different relation tasks in NELL-995 dataset.}
    \label{tb:taskresults}
    \vspace{-10pt}
\end{table}
\paragraph{Link Prediction (Query Answering)}
Given an incomplete KG, the link prediction task aims to predict the missing entities in the unknown links.
In our settings, for a query $(e_1, r, ?)$ or $(?, r, e_2)$, we run multiple rollouts to search for answer node based on query relation and source entity, and then do a beam search with a beam width of 50 to rank the entities by the probability of their trajectories reaching the correct entity. 
Here, we use Hits@$1,3,10$ and Mean Reciprocal Ranking (MRR) as standard ranking metrics \cite{xiong2017deeppath, sutton2018reinforcement}.
\paragraph{Fact Prediction}
Subtly different from link prediction, fact prediction task targets at inferring whether an unknown fact (triple) holds or not.
According to \cite{xiong2017deeppath}, the true test triples are ranked with some generated false triples.
In the experiments, we first remove all links of groundtruth relations in the raw KG. 
Then the dual agents try to infer and walk through the KG to reach the target entity.
Since we share a similar query-answering mechanism as MINERVA \cite{das2017go}, \model~can directly locate the correct entity node for a given query, and eliminate the need to evaluate negative samples any particular relation. 
Note that if \model~fails to reach any of the entities in the set of correct and negative entities, it then falls back to a random ordering of the entities.
Here, we report Mean Average Precision (MAP) scores for various relation tasks of NELL-995 (corresponding to different subsets).

\subsection{Overall Performance}
As demonstrated in Table \ref{tb:mainresults}, 
we first report the performances of \model~ and all baselines on the link prediction task.
The results of NeuralLP are not included on NELL-995 because it can not scale to the size larger dataset. 
We observe that on FB15K-237, the embedding based methods dominate over several neural multi-hop reasoning approaches.
With deeper investigation, we discover that the structural characteristics of FB15K-237 differ significantly from WN18RR and NELL-995, since it contains much larger number of 1-to-M than the M-to-1 relation instances \cite{wan2020reasoning}.
This indicates that the search process of multi-hop reasoning methods is prone to be stuck in the local nodes with high-degree centrality, renders it hard to reach the correct entity.
Note that on WN18RR, neural symbolic methods (MINERVA, NeuralLP, M-Walk) generally beat embedding based methods, with \model~achieving the highest score on Hits@1, 3 and MRR metrics.
On NELL-995, our approach delivers comparable performance to embedding based methods, 
further outperforms MINERVA by a clear margin on all metrics. 
After averaging results on three datasets, we find that our dual-agent based approach leads to overall improvements relative to the single-agent approach with similar settings (MINERVA) by $3.1\%, 2.1\%, 1.7\%, 2.3\%,$ in terms of Hits@1, 3, 10 and MRR.

Table \ref{tb:taskresults} reports the performance of fact prediction on 10 relation tasks of NELL-995.
Our approach produces an encouraging result in most tasks, contributing an average gain of $7.8\%$ relative to the multi-hop neural methods (PRA, DeepPath, MINERVA, and M-walk) and $14.8\%$ gain compared to the embedding-based approaches (TransR and TransE).
\begin{table*}[!t]
\centering
\scalebox{0.68}
{
\begin{tabular}{l l}
\hline \\
(i) \textbf{Can learn shorter path: }
\textsf{Oklahoma Thunder} $\xrightarrow{\text{TeamPlaysInLeague}}$~? \\ \\
\textsf{Oklahoma Thunder} $\xrightarrow{\text{SubPartOfOrganization}}$ 
\textsf{NBA}\\ \\
\textsf{Oklahoma Thunder} $\xrightarrow{\text{SportGameTeam}^{-1}}$ 
\textsf{SportGames} $\xrightarrow{\text{SportGameTeam}}$
\textsf{Boston Celtics} $\xrightarrow{\text{SubPartOfOrganization}}$ 
\textsf{NBA}
\\ \hline \\
(ii) \textbf{Can learn longer path: }
\textsf{Prof. Stephanie Strom} $\xrightarrow{\text{WorkFor}}$~? \\ \\
\textsf{Prof. Stephanie Strom} $\xrightarrow{\text{WritesForPublication}}$ 
\textsf{Times Newspaper} $\xrightarrow{\text{WritesForPublication}^{-1}}$ 
\textsf{Journalist David Johnston} $\xrightarrow{\text{WorksFor}}$ \textsf{NY Times Website}
\\ $\xrightarrow{\text{WritesForPublication}^{-1}}$ 
\textsf{Journalist Adam Liptak} $\xrightarrow{\text{WorksFor}}$
\textsf{Times MusicArtist} \\ 
\hline\\
(iii) \textbf{Can recover from mistakes in longer path: }
\textsf{US Government} 
$\xrightarrow{\text{OrgHiredPerson}}~?$\\ \\
\textsf{US Government} 
$\xrightarrow{\text{AgentParticipatedIn}}$ \textsf{Crime Charge}
$\xrightarrow{\text{AgentParticipatedIn}^{-1}}$
\textsf{Adolescent MusicArtists}
$\xrightarrow{\text{AgentParticipatedIn}}$
\tb{Event Outcome}
\\
$\xrightarrow{\text{AgentParticipatedIn}^{-1}}$
\textsf{"Women of MORE Magazine"} $\xrightarrow{\text{AgentParticipatedIn}}$ 
\tb{Event Outcome}
$\xrightarrow{\text{AgentParticipatedIn}^{-1}}$ 
\textsf{Government Law} $\xrightarrow{\text{AgentParticipatedIn}^{-1}}$ 
\textsf{Ryan Whitney} 
\\ \\ \hline 
\end{tabular}
}
\vspace{-2mm}
\caption{A few examples of paths found by \model~ on NELL-995.}
\label{tab:example_paths}
\vspace{-16pt}
\end{table*}

\begin{figure*}[t]
\setlength{\abovecaptionskip}{0pt}
\centering
\subfloat[AthletePlaysForTeam.]{
\begin{minipage}[t]{0.25\textwidth}
\includegraphics[width=1\textwidth]{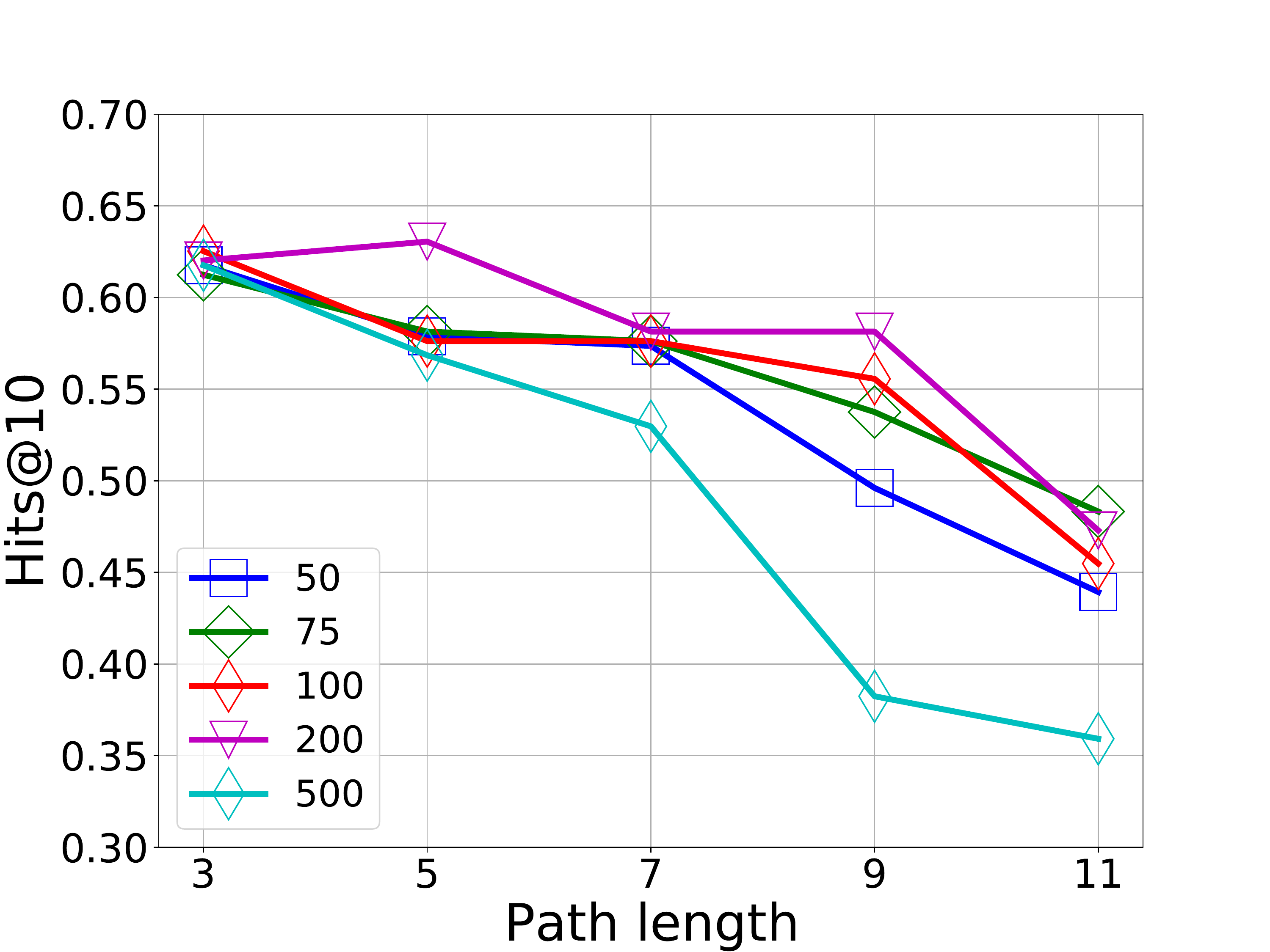}
\vspace{-10pt}
\label{fb}
\end{minipage}%
}
\subfloat[AthletePlaysSport.]{
\begin{minipage}[t]{0.25\textwidth}
\includegraphics[width=1\textwidth]{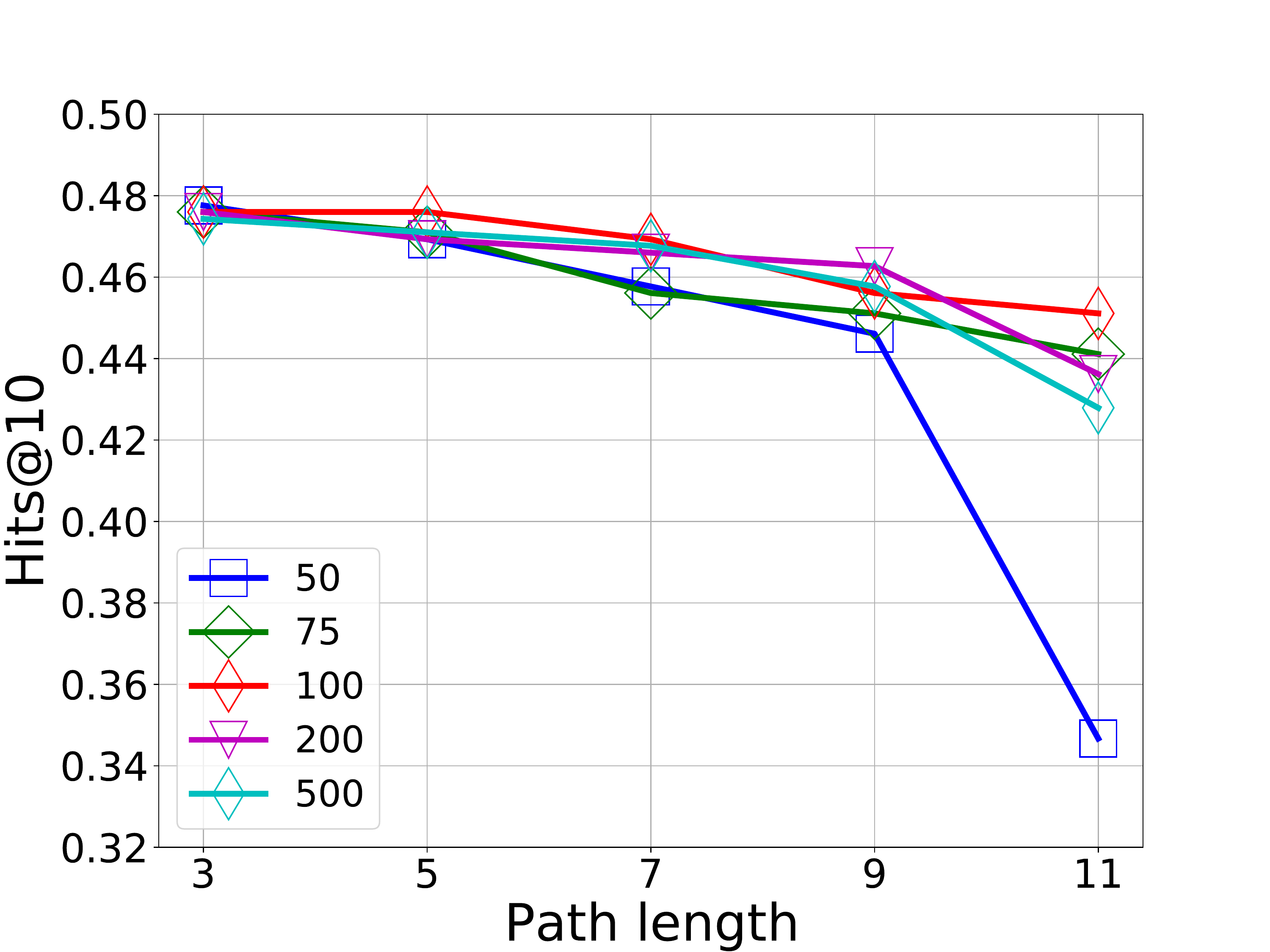}
\vspace{-10pt}
\label{fb}
\end{minipage}%
}
\subfloat[OrgHiredPerson.]{
\begin{minipage}[t]{0.25\textwidth}
\includegraphics[width=1\textwidth]{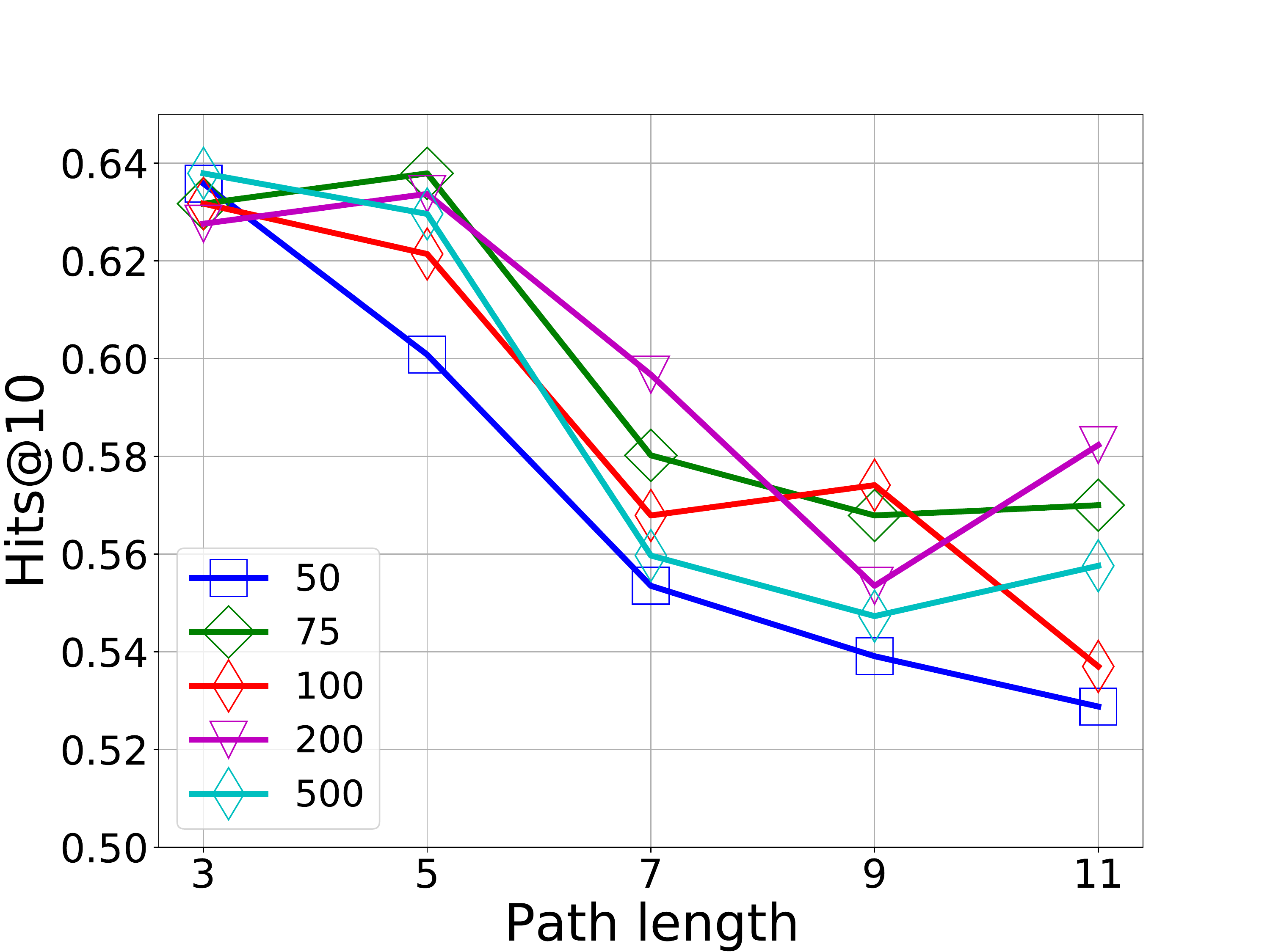}
\vspace{-10pt}
\label{wn}
\end{minipage}%
}
\subfloat[PersonLeadsOrg.]{
\begin{minipage}[t]{0.25\textwidth}
\includegraphics[width=1\textwidth]{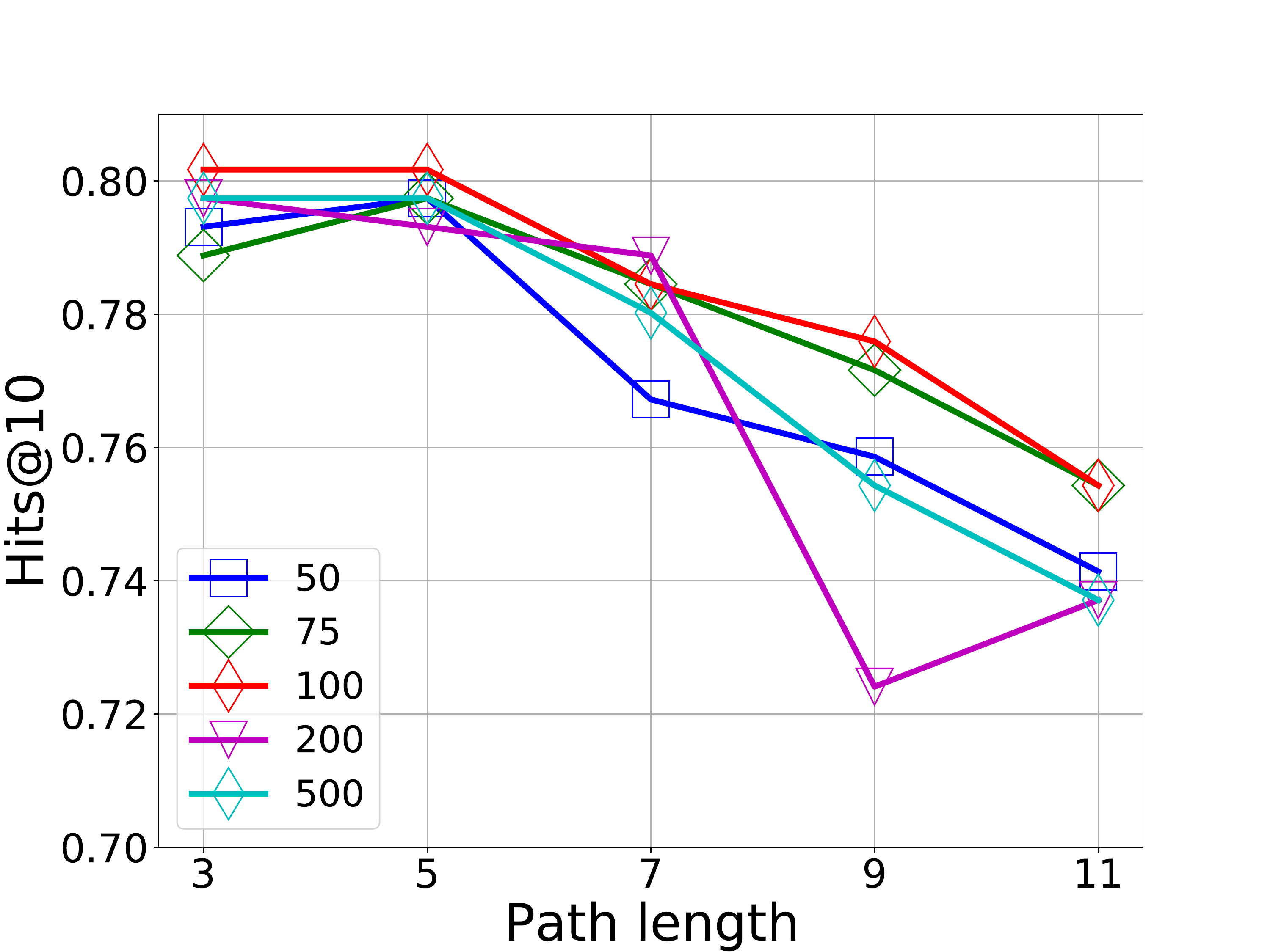}
\label{nell}
\vspace{-10pt}
\end{minipage}%
}
\centering
\vspace{5pt}
\centering
\caption{The effect of different cluster numbers used by \ts{giant}. We present the link prediction performance on four relation tasks on NELL-995. 
Lines in different colors indicate results by different cluster numbers.}
\label{cluster_num_check2}
\vspace{-0.3cm}
\end{figure*}

\begin{figure}[t]
\setlength{\abovecaptionskip}{0pt}
\setlength{\belowcaptionskip}{-20pt}
\centering
\scalebox{0.45}{
\subfloat[OrgHeadqInCity.]{
\begin{minipage}[t]{0.31\textwidth}
\includegraphics[width=1\textwidth]{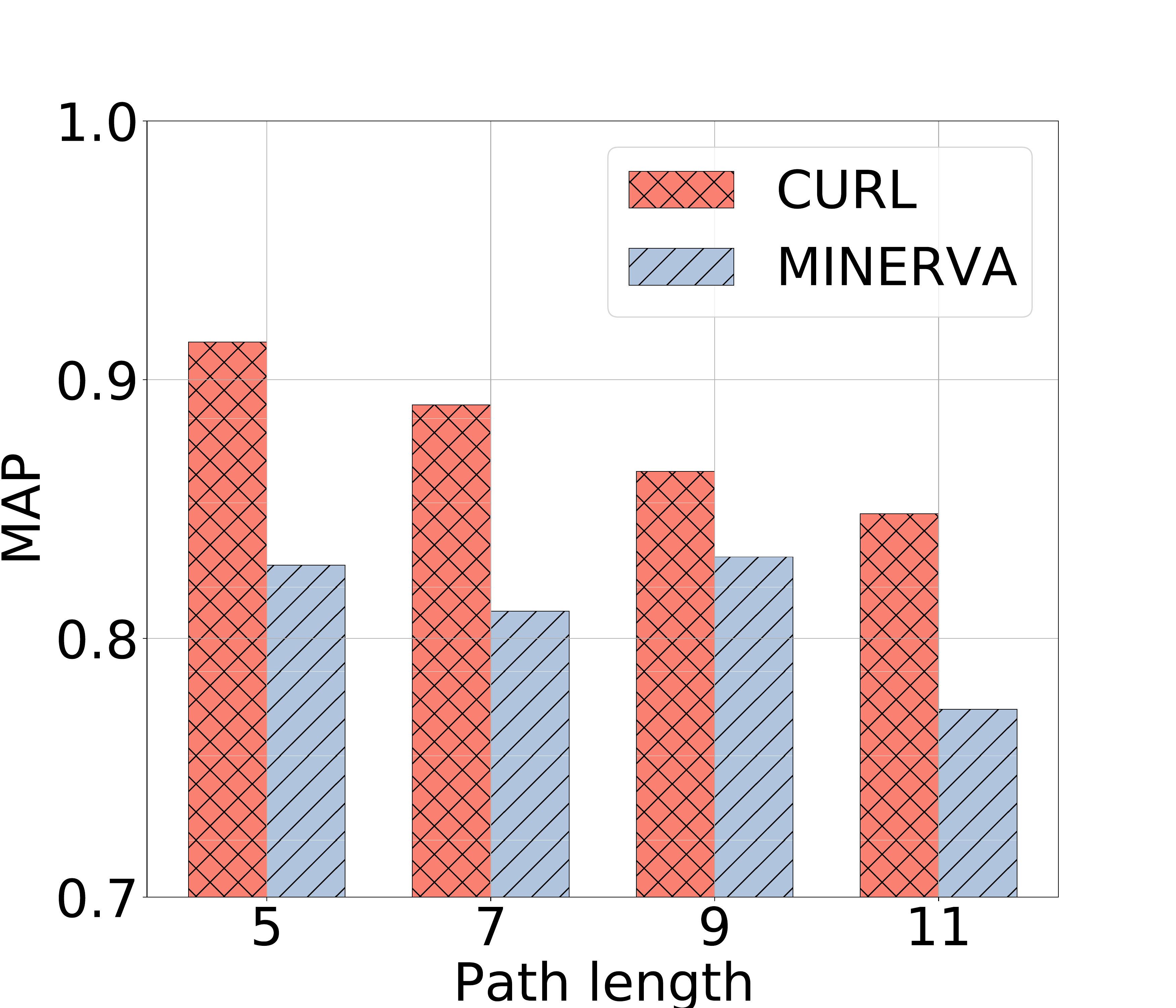}
\vspace{-10pt}
\label{ucncertain}
\end{minipage}%
}
\subfloat[AthletePlaysInLeague.]{
\begin{minipage}[t]{0.31\textwidth}
\includegraphics[width=1\textwidth]{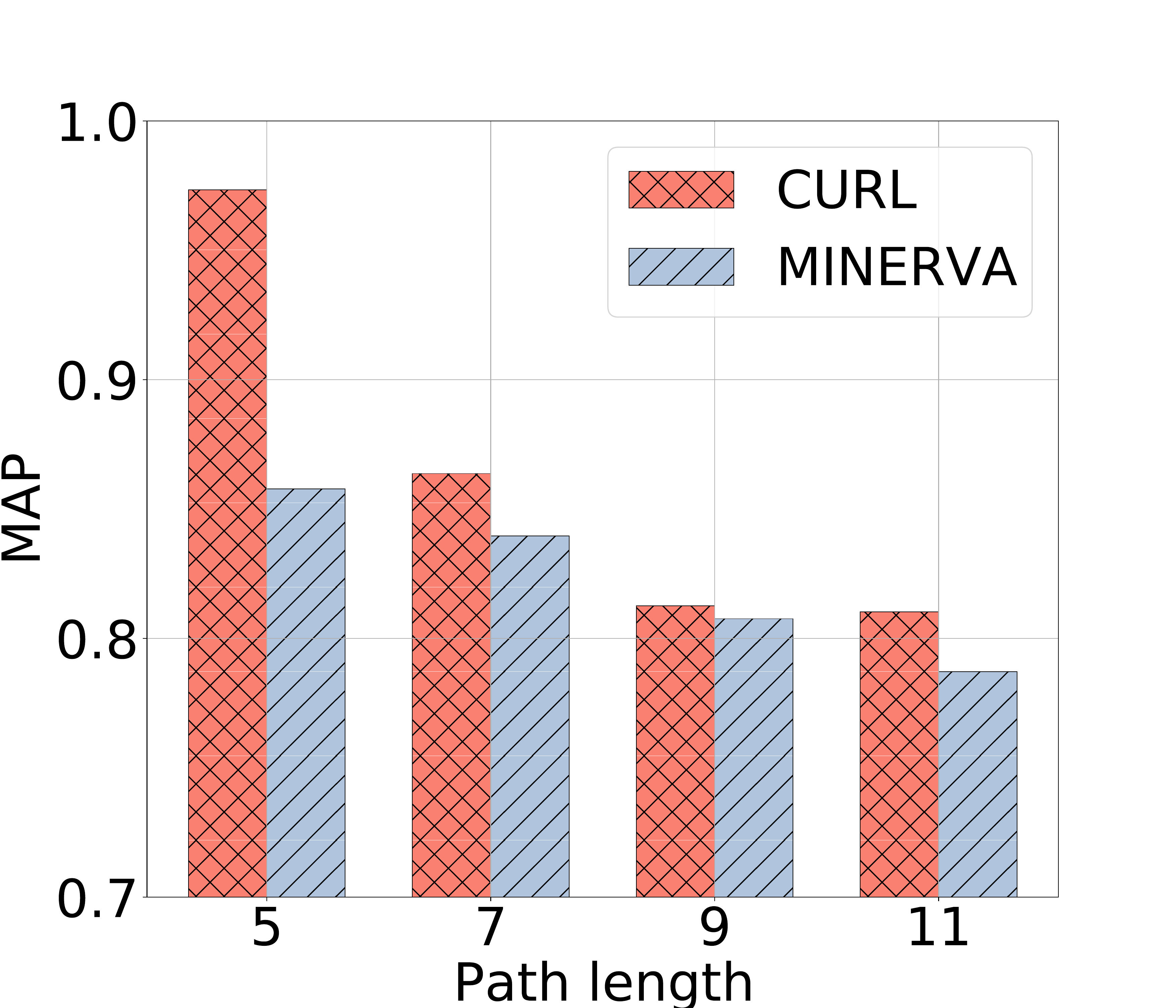}
\vspace{-10pt}
\label{time}

\end{minipage}%
}
\subfloat[TeamPlaysSport.]{
\begin{minipage}[t]{0.31\textwidth}
\includegraphics[width=1\textwidth]{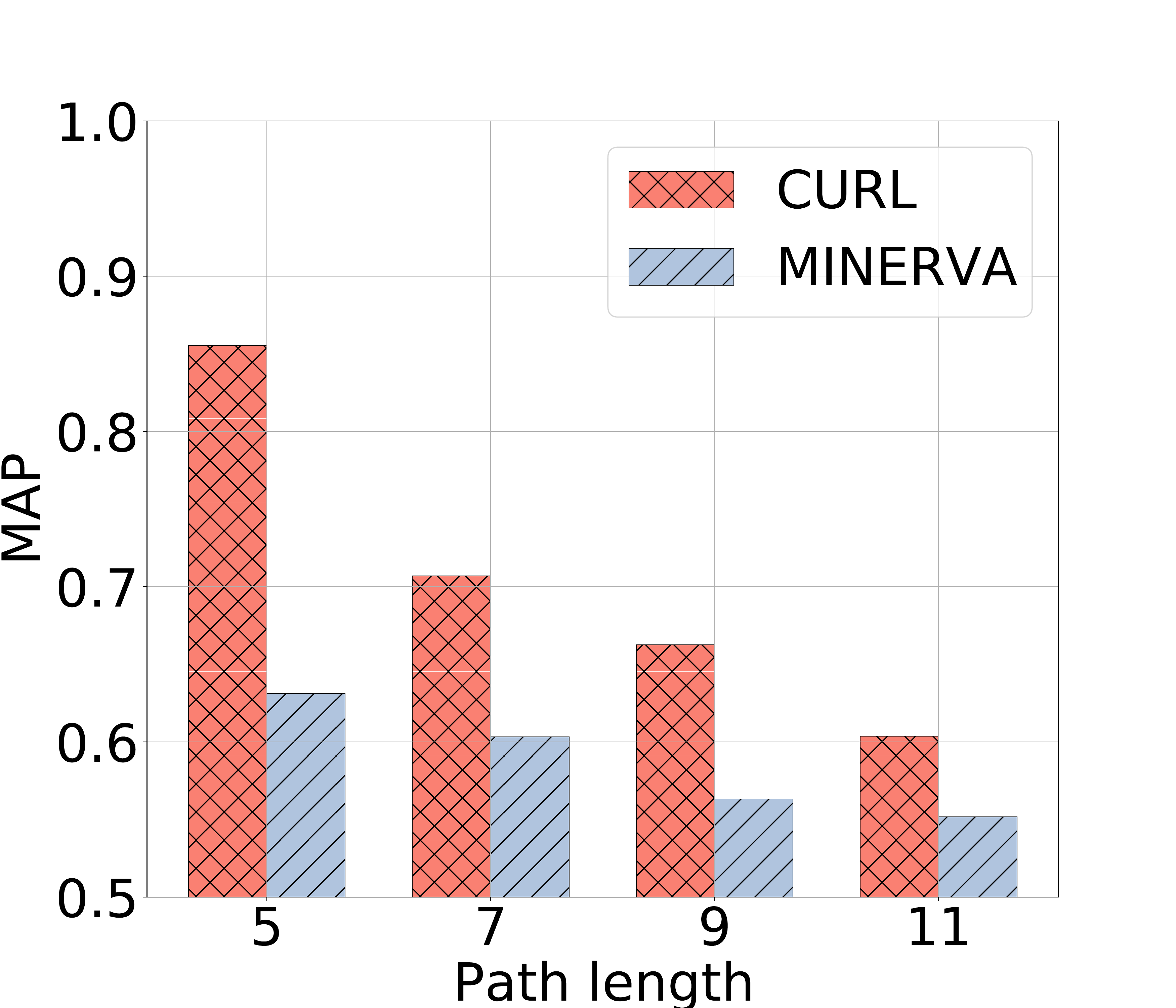}
\vspace{-10pt}
\label{user}
\end{minipage}%
}
}
\vspace{5pt}
\centering
\caption{The long-path performance: \model~significantly outperforms MINERVA on NELL-995.} 
\label{long_paths}
\end{figure}


\subsection{Analysis of \model~and Case Studies}
Based on the above results, we conducted the analysis to discuss the superiority of our dual-agent framework on KG reasoning.
First, 
the above quantitative results show that
our dual-agent design contributes significant gains relative to the \ts{dwarf}-solely method (i.e., MINERVA) on all datasets, 
confirming the effectiveness of our high-level motivation that using cluster-level reasoning to guide entity-level reasoning can alleviate the long path challenge and sparse reward issue.
To further examine this, 
in Figure \ref{positive_reward}, 
we show the positive reward rate (i.e., the percentage of trajectories with positive reward during training) on the NELL-995 tasks. 
Compared to MINERVA under the same training and testing procedure, 
\model~is capable of generating trajectories with more positive rewards, and this continues to improve as training progresses. 
Additionally, 
in Figure 5 (in Appendix B.1),
we show the Hits@1 performance change w.r.t the increasing path length, where we find that \model~maintains a relatively stable result or slower ratio of performance degradation. 
These two evidences prove that introducing \ts{giant} agent by our Collaborative Walking and Mutual Reinforcement Reward can indeed benefit the entity-level KG reasoning.
Finally, we present some illustrative examples of paths found by \model~in Table \ref{tab:example_paths}. 
Example (i) and (ii) illustrate that \model~is capable of capturing both short and long reasoning chains for diverse tasks.
Example (iii) displays the ability to correct a previously taken decision even in the long paths, where our model took an incorrect decision at the first step but was able to revert the decision because of the presence of inverted edges. 
This property is similar to MINERVA, which however cannot well handle the long-path reasoning compared to us. 

\subsection{Long Path Reasoning Performance}
While short chains (length$\leq3$) can partially support KG reasoning and navigate the agent to find target answers \cite{das2017go},
we consider a more practical and rigorous scenario in real-world KGs, where short paths are mostly absent from incomplete KGs.
For effective evaluation, we compare our model with MINERVA in NELL-995 dataset, where we remove the most frequently-visited short paths found by the bi-directional search (BiS) \cite{xiong2017deeppath} inside KG.
In particular, given a triplet $(e_s, r, e_o)$ in the training or testing set, we sample an intermediate entity node $e_i$ and use the breadth-first search algorithm \cite{bundy1984breadth} to verify the traversability from $e_s$ to $e_i$ and from $e_i$ to $e_o$ inside KG. 
After conducting the BiS on each task 50 times, we choose the walkable paths with a length smaller than 3 (self-included), and eliminate their traversed edges inside original KGs. 
Figure \ref{long_paths} plots the MAP scores on varying path lengths in various tasks. 
As can be seen, \model~outperforms MINERVA across all three tasks with different path lengths, and our gains are much more prominent when path length is 5. 
This explains that the cluster-level exploration is essential to lead entity-level agent rather than doing purely random walks in the neighborhood of the source entity. 
Overall, \model~is much more robust to queries where a longer reasoning path is required, showing minimal degradation in performance for even the longest path setting.
Additional experiments on FB15K-237, WN18RR, and NELL-995 can be found in Appendix B.2 of the supplementary material.

\subsection{Sensitivity Test on Cluster Size}

Figure \ref{cluster_num_check2} shows the Hits@10 scores of \model~under different cluster number $N$ during training on four NELL-995 tasks.
We observe a performance improvement when we increase $N$ from $50$ to $75$ and performance degrade when we further increase $N$ from $75$ to $500$. 
These results illustrate $75$ cluster number is powerful enough to capture high-level semantic information.
In other words, parametrized by a suitable $N$, our approach is generally stable when the path lengths increase, indicating that proper clustering of KG entities does refine a series of meaningful high-level semantics to facilitate the low-level path searching. 


\section{Related Work}

\paragraph{Knowledge Graph Representation Learning} 
Representation learning has shown great success in a wide range of fields \cite{zhang2019job2vec, zhang2017efficient, li2021interpretable, yuan2021self} and KG reasoning is no exception.
Recent advances in this area have proposed a variety of embedding-based methods that project the entities and relations into low-dimensional continuous vector space by exploiting entity types \cite{guo2015semantically, ouyang2017representation}, relation paths \cite{lin2015modeling, toutanova2016compositional, li2018link, zhang2018discriminative}, textual descriptions \cite{zhong2015aligning, xiao2017ssp}, and logical rules \cite{omran2018scalable, hamilton2018embedding}.
For instance, TransE \cite{wang2014knowledge} first encoded the entities and relations into latent vectors by following translational principle in point-wise Euclidean space. 
Besides, ComplEx \cite{trouillon2016complex} introduced complex vector space to capture both symmetric and anti-symmetric relations.
However, such models ignored the symbolic compositionality of KG relations, making them unable to discover complex reasoning paths with one-hop distance-based measure \cite{wang2017knowledge, ji2021survey}. 
Furthermore, it is hard to interpret the traversal paths, and these models can be computationally expensive to access the entire graph in memory.

\paragraph{Deep Reinforcement Learning for KG Reasoning}

The emergence of deep reinforcement learning (RL) enables many path-based approaches to learn symbolic inference rules from relational paths inside KG \cite{ji2021survey}.
By formulating KG reasoning as a sequential decision problem and taking multi-hop random walks, existing studies improve the empirical performance of various tasks \cite{xiong2017deeppath, shen2018m, lin2018multi, das2017go, wan2020reasoning, hildebrandt2020reasoning} including KG completion, fact prediction, and query answering.
Specifically, DeepPath \cite{xiong2017deeppath} first introduced RL to search for diversified representative paths between entity pairs.
M-Walk \cite{shen2018m} further proposed to solve the reward sparsity problem in MCTS-based query answering in an off-policy manner.
Reward shaping based approach \cite{lin2018multi} leveraged pre-trained embeddings to estimate the likelihood of unseen facts, 
with the purpose of reducing the impact of false-negative supervision as well as facilitating the path inference.
Our work aligns with the RL formulation of MINERVA \cite{das2017go}, i.e., learning to walk and search for answer entities of a particular KG query in an end-to-end fashion.
Note that MINERVA solely exploited the entity-level KG information to update the policy network. 
In comparison, we trained the entity-level policy of the \ts{Dwarf Agent} using the cluster-level semantics and trajectories generated by the \ts{Giant Agent}, such that the dual agents can collaborative to reach the optimal answers given queries.

Regarding the dual-agent structure, our work is closely related to the hierarchical RL~\cite{barto2003recent, wang2018deep, li2018path, wan2020reasoning, wang2020h2kgat, wen2020efficiency} in the sense of leveraging both low-level and high-level policies to solve long-horizon problems with sparse rewards.
However, instead of requiring high-level policy to specify subgoals for low-level skills, \model~enables both policies to achieve the subgoals possessed by their counterparts. 
That is, with the faster convergence in a reduced search space, \ts{giant agent} can provide high-level stepwise guidance for \ts{dwarf agent} to follow, while \ts{dwarf agent} walks along the real-existing paths, thus checking the correctness of abstractive cluster-level paths found by \ts{giant agent}. 
Such a learning scheme can avoid extra design efforts for complex subgoal space, which is not always trustworthy and tractable~\cite{zhang2019interstellar}.




\section{Conclusion}
We proposed a dual-agent framework (\tb{\model}) that learns to walk over a KG for searching desired answer nodes given query relation and source entity.
Specifically, we first leveraged LSTM to project trajectory history into latent vectors of different granularities and semantics.
To facilitate the entity-level exploration, \model~launched two agents: \ts{giant} and \ts{dwarf} to collaboratively explore paths at different granularity levels and search for the answer.
\ts{Giant} walks rapidly over inner clusters of the KG, which can guide \ts{dwarf} to smoothly traverse through the entities inside the clusters.
We later developed the collaborative policy network for sharing historical path information between two agents, and established the mutual reinforcement reward system for handling sparse reward issue.
Experimental results on several knowledge graph reasoning benchmarks show that our approach can search for answers more accurately and efficiently.
Furthermore, we compared with the \ts{dwarf}-solely MINERVA in the long-path experiment.
We found that our method is more accurate in long path reasoning, which can be explained by that the stage-wise signals provided by \ts{giant} do play a critical role in leading the \ts{dwarf} towards the target node.

\section{Acknowledgements}
This work was partially supported by the National Science Foundation through award 2006387, 1814510, 2040799.

\bibliography{ref}

\clearpage
\newpage
\appendix
\section{Appendix}
\section{A. Experiment Details}
\label{Appendix:exp_details}

\paragraph{A.1. Statistics of Datasets}

The NELL-995 knowledge dataset contains $75,492$ unique entities and $200$ relations. 
WN18RR contains $93,003$ triples with $40,943$ entities and $11$ relations. And FB15K-237, a subset of FB15K where inverse relations are removed, contains $14,541$ entities and $237$ relations. The detailed statistics are shown in Table \ref{tab:data_stats}.
\begin{table}[H]
\begin{center}
\huge
\scalebox{0.46}{
\begin{tabular}{c c c c c c c}\hline
\multirow{2}{*}{\textbf{Dataset}} & \multirow{2}{*}{\textbf{\#entities}} & \multirow{2}{*}{\textbf{\#relations}} & \multirow{2}{*}{\textbf{\#facts}} & \multirow{2}{*}{\textbf{\#queries}} & \multicolumn{2}{c}{\textbf{\#degree}} \\
&&&&& \textbf{mean} & \textbf{median} \\\hline
\textsc{WN18RR} & 40,945 & 11 & 86,835 & 3,134 & 2.19 & 2\\
\textsc{NELL-995} & 75,492 & 200 & 154,213 & 3,992 & 4.07 & 1 \\
\textsc{FB15k-237} & 14,505 & 237 & 272,115 & 20,466 & 19.74 & 14\\\hline
\end{tabular}
}
\caption{Statistics of three datasets used in our experiments.}
\label{tab:data_stats}
\end{center}
\end{table}

\paragraph{A.2. Optimal Hyperparameters}
\label{Appendix:exp_details}
To reproduce the results of our model in Table \ref{tb:mainresults} and Table \ref{tb:taskresults}, we report the empirically optimal hyperparameters.
Specifically, we set the entity embedding dimension to $50$ and relation embedding dimension to $50$.
We use the K-means algorithm \cite{macqueen1967some} and the pre-trained entity embeddings to initialize $75$ clusters for WN18RR and NELL-995, and $50$ clusters for FB15K-237, respectively.
After the maximum step $T$ has been reached, \model~evaluates the action sequence and assigns the mutual rewards according to lines $10$, $11$ in Algorithm $1$.
The action embedding is the concatenation of the entity embedding vector and the relation embedding vector. 
We use two single-layer LSTMs with hidden state size of $200$ for \ts{giant} and \ts{dwarf}, respectively. 
On all datasets, the quantities of path rollouts in training and testing are $20$ and $100$, separately.
We use the Adam optimization \cite{kingma2014adam} in REINFORCE for model training with learning rate as $0.001$, and the best mini-batch size is $128$.
For the rest parameters, i.e.,
maximum path length $T$, the entropy regularization constant $\beta$, and the moving average constant $\lambda$,
the best combination of them are $\{3, 0.2, 0.2\}$ for FB15K-237, $\{3, 0.06, 0.00\}$ for WN18RR, $\{3, 0.07, 0.07\}$ for NELL-995.

\section{B. Additional Experiments}
\label{Appendix:AdditionalExperiments}

\begin{figure}[h]
\setlength{\abovecaptionskip}{0pt}
\setlength{\belowcaptionskip}{-3pt}
\centering
\subfloat[FB15K-237.]{
\begin{minipage}[t]{0.16\textwidth}
\includegraphics[width=1\textwidth]{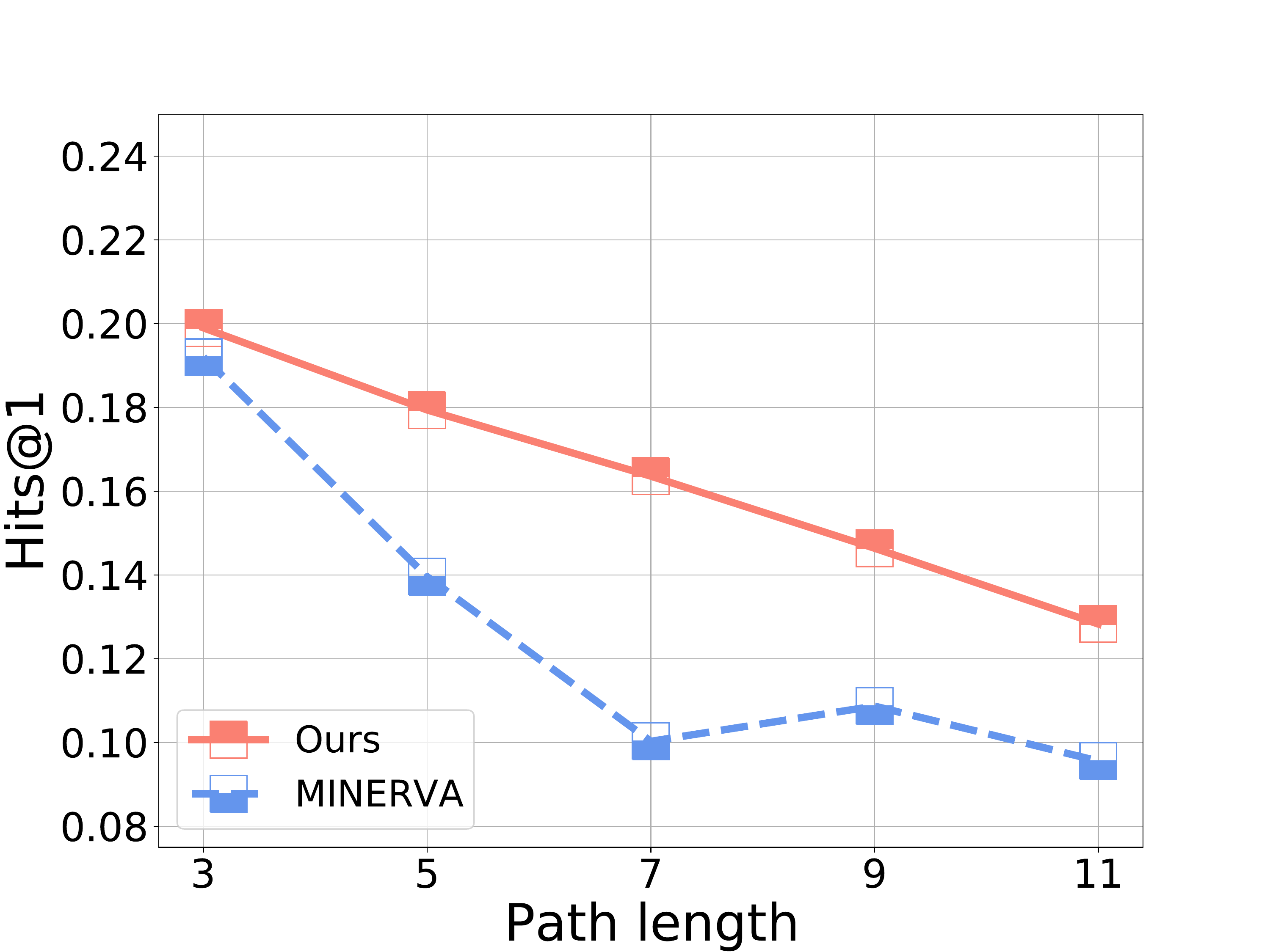}
\vspace{-10pt}
\label{fb}
\end{minipage}%
}
\subfloat[WNRR18.]{
\begin{minipage}[t]{0.16\textwidth}
\includegraphics[width=1\textwidth]{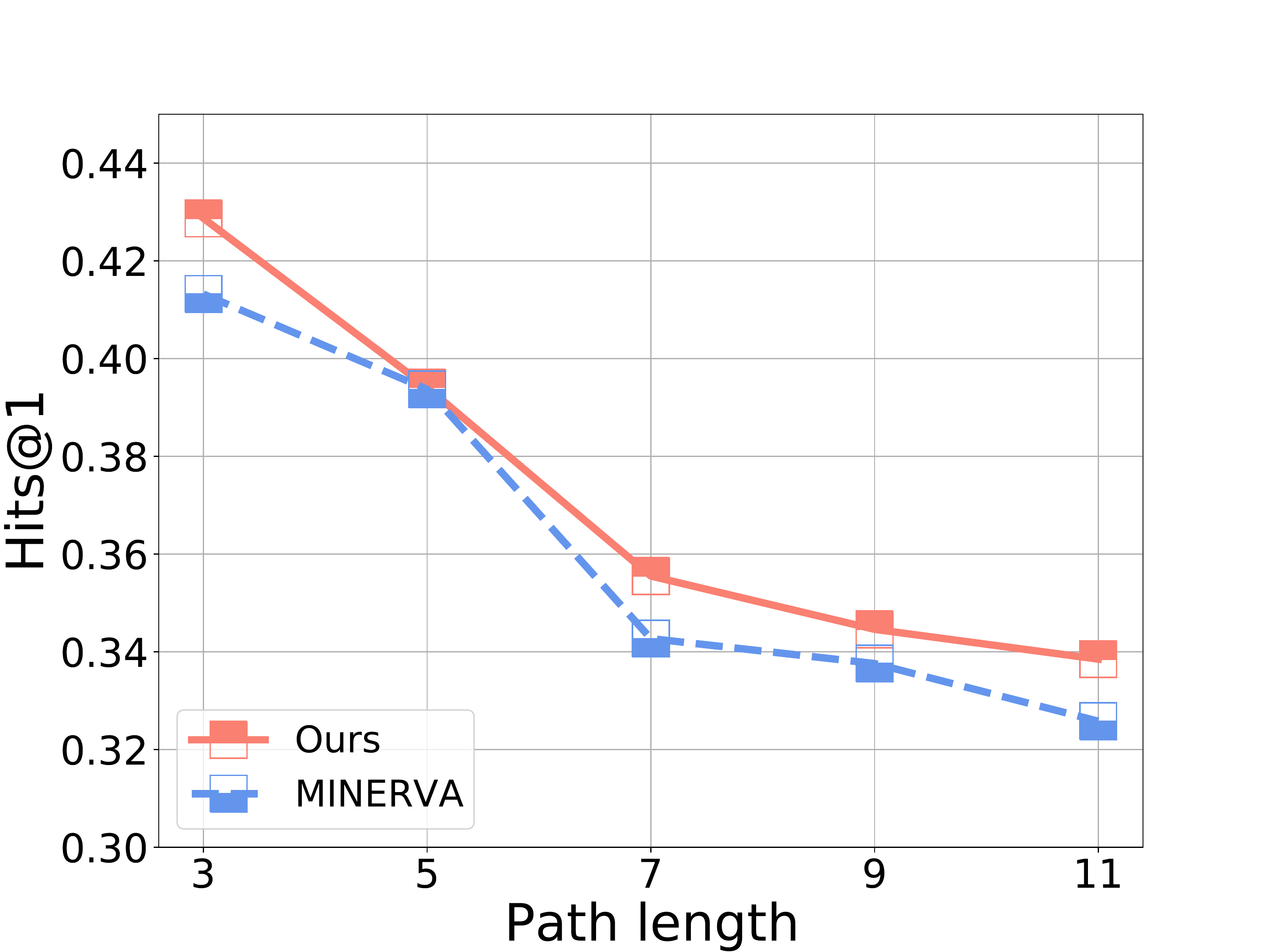}
\vspace{-10pt}
\label{wn}
\end{minipage}%
}
\subfloat[NELL-995.]{
\begin{minipage}[t]{0.16\textwidth}
\includegraphics[width=1\textwidth]{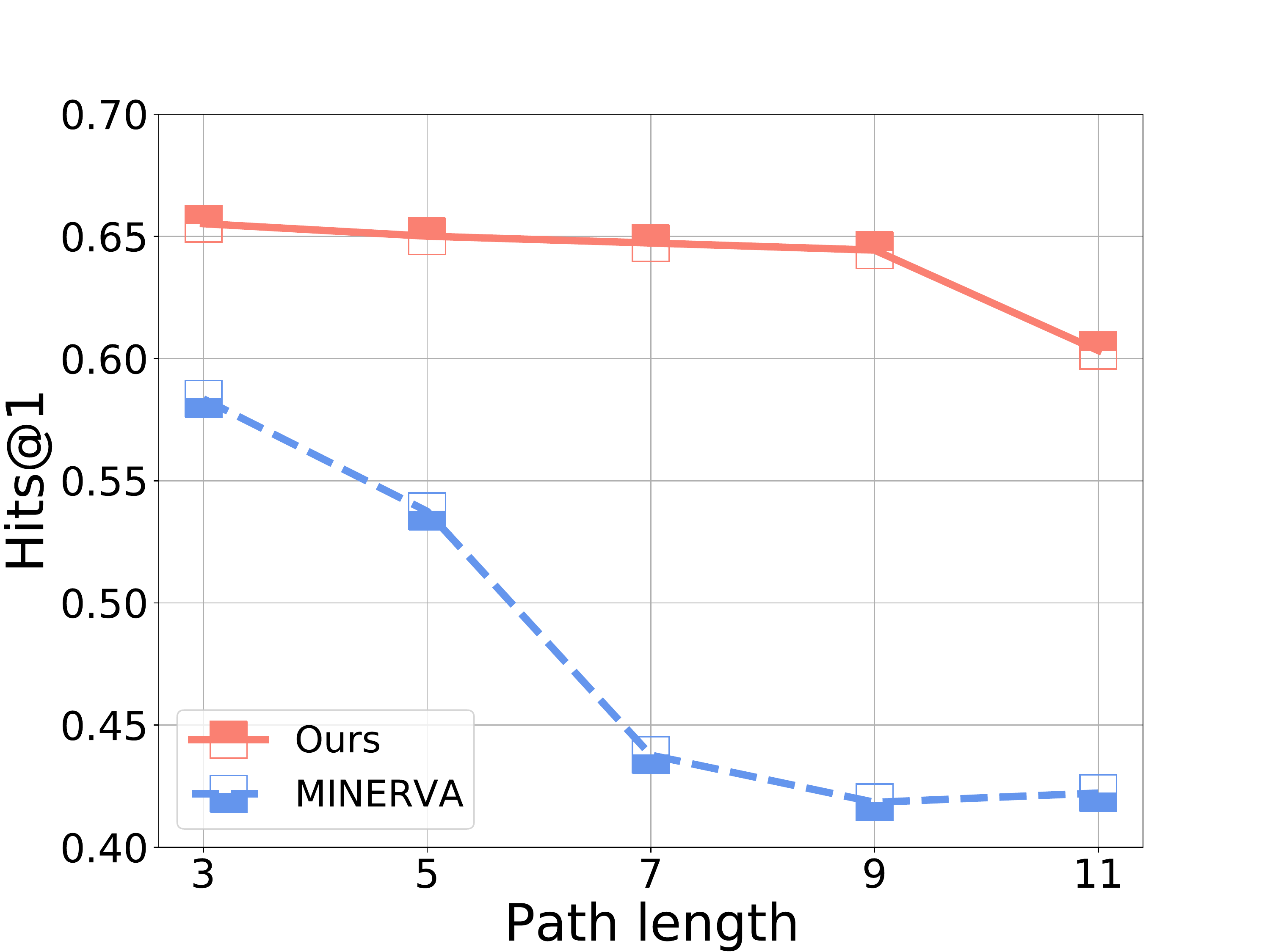}
\label{nell}
\vspace{-10pt}
\end{minipage}%
}
\centering
\vspace{3pt}
\caption{The long-path experiments: we significantly outperform MINERVA in three large datasets.}
\label{additional_long_path_experi}
\end{figure}

\begin{table}[th]
    \centering
    \small
    \setlength{\abovecaptionskip}{10pt}
    \scalebox{0.9}{
    \begin{tabularx}{0.44\textwidth}{m{1cm}<{\centering}|m{1.4cm}<{\centering}|m{2cm}<{\centering}m{2cm}<{\centering}}
    
        \hline
    	\multicolumn{2}{c|}{\textbf{Size}} & MINERVA & CURL \\ \hline
    	\multirow{3}{*}{20} & Hits@1 & 72.3 & \bf{74.7} \\
    	& Hits@10 & \bf{81.9} & 81.1\\ 
    	& MRR & 75.8 & \bf{77.2}\\ \hline
    	
    	\multirow{3}{*}{50} & Hits@1 & 70.7 & \bf{75.9} \\
    	& Hits@10 & 81.1 & \bf{82.3}\\ 
    	& MRR & 74.9 & \bf{78.5}\\ \hline
    	
    	\multirow{3}{*}{100} & Hits@1 & 59.4 & \bf{76.3} \\
    	& Hits@10 & 76.3 & \bf{82.5}\\ 
    	& MRR & 66.0 & \bf{78.9}\\ \hline
    	
    	\multirow{3}{*}{200} & Hits@1 & 69.9 & \bf{75.1} \\
    	& Hits@10 & 79.9 & \bf{82.7}\\ 
    	& MRR & 73.6 & \bf{78.2}\\ \hline
    	
    	\multirow{3}{*}{500} & Hits@1 & 68.7 & \bf{73.1} \\
    	& Hits@10 & 78.3 & \bf{83.9}\\ 
    	& MRR & 72.0 & \bf{77.1}\\ \hline
    	
    \end{tabularx}
    }
     \caption{Hits@1, 10 and MRR test accuracy (\%) in task of OrganizationHeadquarteredInCity, where ``Size'' denotes the width of beam search. 
    \label{tb:beam_size}
    }
\end{table}

\subsection{B.1. Parameter Sensitivity on Beam Search Size}
\label{Appendix:ecn}

We also check the influence of different beam search size during testing, as reported in Table \ref{tb:beam_size}.
As can be seen, except when beam size is 20, the performances of \model~are much higher than MINERVA, and our gain is maximized at beam search size 50. 
Moreover, the test accuracy doesn't change substantially with larger beam search sizes.

\subsection{B.2. Long Path Recovery}
\label{Appendix:lpt}
We conduct further long-path experiments on three datasets, including FB15K-237, WN18RR, and NELL-995.
Here, we investigate the ability of recovering from mistakes in the long-path traversal.
Unlike Section "Long Path Reasoning Performance", we retain their original KGs and set the walkable path length to be relatively larger ($> 3$), since short chains in these datasets can usually produce good empirical results \cite{das2017go, wan2020reasoning}, and the longer ones are more likely to walk through unnecessary or incorrect links.
As reported in Figure \ref{additional_long_path_experi}, \model~outperforms the other \ts{Dwarf}-only method (i.e., MINERVA) in all datasets. 
This observation demonstrates that dual-agent design is able to stabilize the entity-level searching in long paths by utilizing the cluster-level reasoning information.
One possible reason is that \ts{Giant} converges much faster in a reduced search space, consistently providing high-level stage-wise guidance for \ts{Dwarf} to follow.

\end{document}